%% file: example.tex
\documentclass{article}

\usepackage[preprint]{corl_2026} 

\usepackage{amsmath}
\usepackage{graphicx}
\usepackage{booktabs}
\usepackage{multirow}
\usepackage{pifont}
\usepackage{makecell}
\usepackage{enumitem}
\usepackage{xspace}
\usepackage{algorithm}
\usepackage{algorithmic}
\usepackage{caption}
\usepackage{stfloats}
\usepackage{lineno}
\definecolor{darkgreen}{rgb}{0.17,0.56,0.36}

\newcommand{\eg}{\emph{e.g.}}
\newcommand{\ie}{\emph{i.e.}}
\newcommand{\correct}{\textcolor{darkgreen}{\ding{51}}}
\newcommand{\wrong}{\textcolor{red}{\ding{55}}}
\newcommand{\proposed}{SFHand\xspace}
\newcommand{\dataset}{EgoHaFL\xspace}

\newcommand{\rev}[1]{\textcolor[rgb]{0.0,0.0,0.0}{#1}}

\title{Predictive Representation Learning for Manipulation via Streaming Egocentric 3D Hand Forecasting}

\author{
  Ruicong Liu, Yifei Huang\thanks{Corresponding author.}, Liangyang Ouyang, Caixin Kang, Yoichi Sato\\
  The University of Tokyo, Japan\\
  \texttt{\{lruicong,hyf,oyly,cxkang,ysato\}@iis.u-tokyo.ac.jp} \\
}

\begin{document}
\maketitle


\input{sec/0_abstract}
\input{sec/1_intro}
\input{sec/2_related}
\input{sec/3_method}
\input{sec/4_dataset}
\input{sec/5_experiment}
\input{sec/6_conclusion}
\bibliography{example}

\input{sec/X_suppl}

\end{document}

%% file: sec/0_abstract.tex
\begin{abstract}
  Learning embodied manipulation from egocentric human data offers a scalable path, reducing the cost of collecting robot-specific demonstrations. However, existing representation learning methods largely stop at perceptual understanding, such as extracting what a scene contains or what an action is. A capable embodied agent is more than understanding: it predicts the next motion. We propose that forecasting 3D human hand motion is one suitable pretraining task, as the representation jointly learns how to manipulate objects and how the motion will unfold next.
  Building on this hypothesis, we introduce \proposed, the first representation learning method via streaming egocentric 3D hand forecasting.
  \proposed learns by autoregressively predicting future 3D hand states, including hand type, 2D bounding box, 3D pose, and trajectory, from a continuous stream of video and language instructions.
  Its autoregressive architecture is equipped with an ROI-enhanced memory layer, capturing temporal context while focusing on salient hand-centric regions.
  To enable this research, we also introduce \dataset, the first open-source large-scale dataset pairing synchronized 3D hand poses with fine-grained language instructions (247K clips, 3.95M frames).
  Transferred to downstream embodied control, the learned representation shows state-of-the-art performance on both simulation and real-world manipulation benchmarks, improving success rate by up to 13.4\%. Our method also achieves leading 3D hand forecasting accuracy, with 35.8\% lower ADE than baseline methods.
  These results suggest that forecasting hand motion is a better pretraining task for representation learning in embodied manipulation. Code: \href{https://github.com/ut-vision/SFHand}{ut-vision/SFHand}.
\end{abstract}

\keywords{Representation learning, Egocentric hand forecasting, Manipulation}

%% file: sec/1_intro.tex
\section{Introduction}
\label{sec:intro}

Embodied AI has shown promise across various manipulation tasks~\citep{R:chen2024end,R:wu2023policy,R:yang2024visual}, yet its progress is limited by data scarcity: collecting large-scale robot demonstrations is expensive and embodiment-specific.
A scalable alternative is learning representations from egocentric human data~\citep{R:nair2023r3m,R:karamcheti2023language,R:ma2023liv,R:radosavovic2023real,R:zeng2024learning,I:huang2024egoexolearn}. Egocentric videos naturally capture human-object interactions, making it a rich source for learning manipulation-relevant representations. Previous representation learning approaches pretrain on those videos via contrastive learning~\citep{R:sermanet2018time,R:ma2022vip}, masked auto-encoder~\citep{R:radosavovic2023real}, or frame reconstruction~\citep{R:zeng2024learning}. However, these objectives share a common limitation: they only encode what a scene contains or what an action is, stopping at perceptual understanding. Instead, a capable embodied agent does more than perceive: it predicts the next motion, an ability that remains unaddressed.

\begin{figure*}[t]
  \centering
  \includegraphics[width=.95\linewidth]{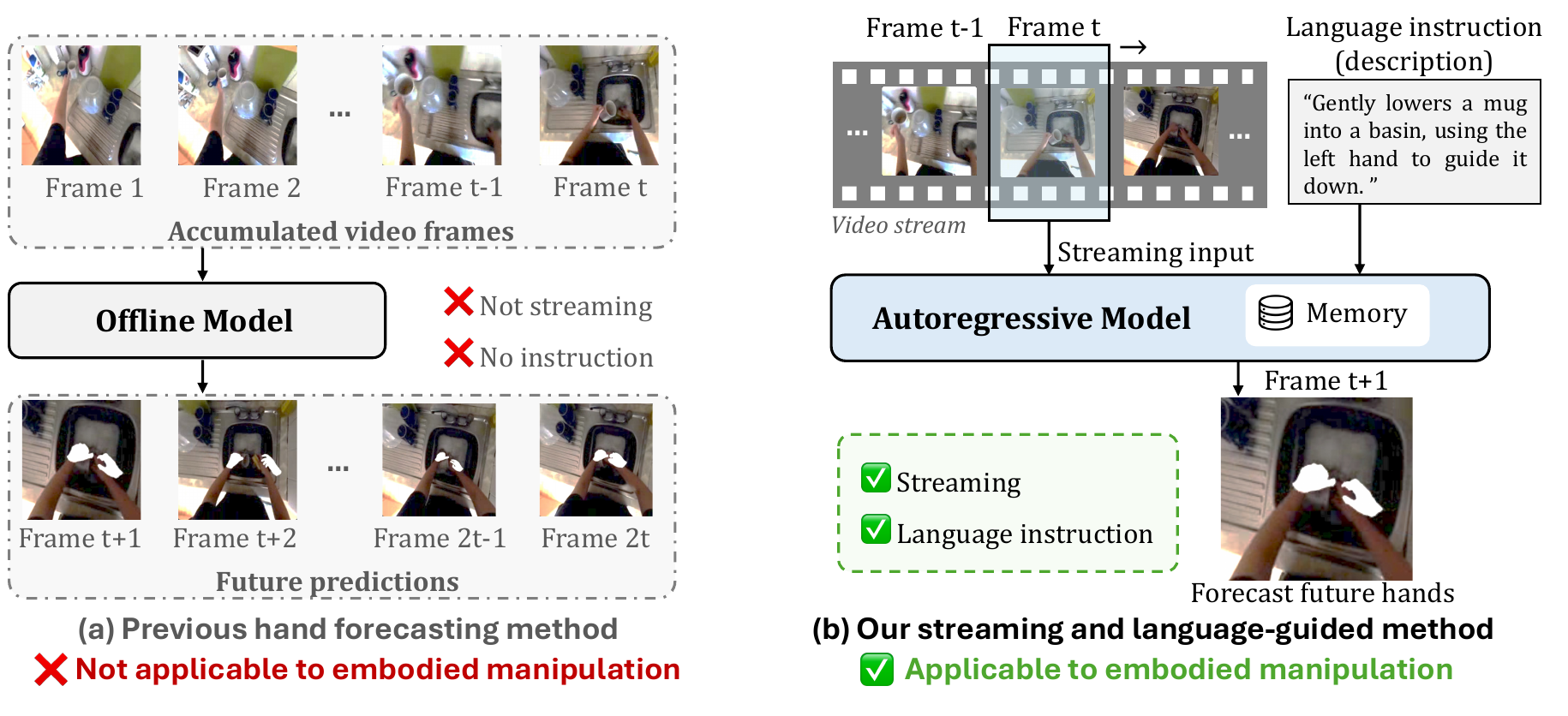}
  \vspace{-.5em}
  \caption{Comparison between previous hand forecasting methods and our proposed approach. (a) Prior methods rely on accumulated video sequences and lack streaming input or language guidance. (b) Our method, \proposed, is an autoregressive framework for streaming language-guided 3D hand forecasting, making it well-suited as a representation learning objective for embodied manipulation.}
  \label{fig:teaser}
  \vspace{-1.5em}
\end{figure*}

In this paper, we argue that \emph{forecasting 3D human hand motion} is one suitable pretraining objective for representation learning. Forecasting hand motions forces a model to jointly learn two things: how humans manipulate objects, and how the hand motion will unfold next. These are precisely the two abilities embodied control requires.
Unfortunately, existing 3D hand forecasting methods are poor candidates for such representation learning. As illustrated in \cref{fig:teaser}(a), recent diffusion-based~\citep{H:hatano2025invisible} and transformer-based~\citep{H:bao2023uncertainty} methods suffer from two fundamental limitations. First, they are \emph{offline} design, requiring the accumulation of multiple observation frames before making a prediction.
This precludes their use in any real-time setting.
Second, they are \emph{unimodal}, relying solely on visual inputs and unable to incorporate natural-language instructions. These shortcomings make existing forecasting methods ill-suited to the demands of embodied control.

To address these challenges, we propose \proposed, the first representation learning method via streaming egocentric 3D hand forecasting. As shown in \cref{fig:teaser}(b), \proposed autoregressively forecasts future 3D hand states (pose, trajectory, type, bounding box) while continuously processing streaming video inputs and a textual instruction. This streaming architecture eliminates the dependence on accumulated observations, allowing the model to be used as a real-time encoder in downstream control loops. To support temporal reasoning over the streaming input, we introduce an ROI-enhanced memory layer that efficiently retains and updates salient information from hand-centric regions, providing the temporal context essential for both forecasting and downstream manipulation.

A key barrier to training such forecasting models is the lack of large-scale, annotated data. To this end, we construct \dataset (Egocentric 3D Hand Forecasting with Language), the first open-source large-scale egocentric dataset that pairs synchronized, sentence-level natural-language descriptions with precise 3D hand annotations. With 247K clips and 3.95M annotated frames, \dataset enables, for the first time, training streaming and language-guided 3D hand forecasting at scale.

We evaluate the transferability of the learned representations. Following standard practice~\citep{R:nair2023r3m,R:karamcheti2023language,R:radosavovic2023real}, we freeze the pretrained encoders and attach lightweight task-specific policy heads on top. \proposed achieves state-of-the-art success rates on both the gripper-based Franka Kitchen~\citep{R:gupta2020relay} and the dexterous-hand Adroit~\citep{R:rajeswaran2017learning} benchmarks, confirming that forecasting-derived features transfer effectively across distinct robot embodiments. On the forecasting task itself, \proposed also shows leading accuracy on \dataset. These results suggest that forecasting hand motion is a better pretraining task for representation learning in embodied manipulation.
Our contributions are as follows:

\begin{itemize}[leftmargin=1.5em]
  \item We propose 3D hand forecasting as a representation learning objective that learns both manipulation and predictive knowledge.
  \item We introduce \proposed, the first streaming, language-guided framework realizing this objective, built around an autoregressive architecture with an ROI-enhanced memory layer.
  \item We build \dataset, the first open-source large-scale dataset pairing synchronized 3D hand poses with fine-grained language instructions. Evaluations show that representations pretrained on it achieve state-of-the-art performance on both 3D hand forecasting and downstream manipulation.
\end{itemize}

%% file: sec/2_related.tex
\section{Related work} \label{sec:related}
\subsection{Embodied representation learning}
Learning robust visual representations is a cornerstone for enabling robots to perceive and interact with complex environments \citep{R:burns2023makes,R:ling2026blue,R:chen2024end,R:hansen2022pre,R:wu2023policy,R:yang2024visual,R:yang2026judge}. As collecting large-scale, task-specific robotic data remains challenging, many recent efforts have explored transferring representation from pre-trained visual models \citep{R:zeng2024learning,R:radosavovic2023real}. Models such as ImageNet-based \citep{N:deng2009imagenet} CNNs and CLIP-based \citep{N:radford2021learning} vision–language encoders have shown notable gains in robotic policy learning \citep{R:khandelwal2022simple,R:parisi2022unsurprising,R:shridhar2022cliport}.
More recently, works has shifted to using egocentric video datasets~\citep{E:grauman2022ego4d,E:damen2018scaling}, which better capture the viewpoint and dynamics of embodied object manipulation.
Various pre-training strategies, such as time-contrastive learning \citep{R:sermanet2018time,R:ma2022vip} and masked modeling \citep{R:radosavovic2023real}, have been introduced to learn temporally consistent and spatially grounded representations.
In parallel, frameworks like R3M \citep{R:nair2023r3m}, Voltron \citep{R:karamcheti2023language}, LIV \citep{R:ma2023liv}, and MPI \citep{R:zeng2024learning} integrate language to enhance semantic grounding and instruction following. However, these methods primarily rely on 2D-based objectives (\ie, contrastive loss or frame prediction) and lack an explicit understanding of 3D structure or temporal dynamics, limiting their ability to model continuous and causal human motion. Our work bridges this gap. By training a streaming, language-conditioned model to explicitly forecast 3D hand states, \proposed learns representations that capture the causal, physical nature of human motion, which we show transfers powerfully to embodied manipulation.

\subsection{Hand forecasting}
Early studies on hand motion forecasting centered on predicting 2D hand trajectories from egocentric videos to infer human intentions and interactions \citep{H:liu2020forecasting,I:huang2020improving,H:liu2022joint,I:huang2025vinci}. These methods often couple trajectory forecasting with action anticipation or interaction hotspot prediction, leveraging hand motion as a critical cue for understanding upcoming actions. Later efforts incorporate ego-motion cues such as head or camera movement \citep{H:ma2024diff,H:ma2024madiff,H:hatano2024emag,I:huang2022compound,I:yang2023deco,I:huang2023weakly}, which improved spatial coherence but remained constrained to the image plane. This 2D-only prediction limits physical realism and spatial reasoning, as real-world interaction is inherently three-dimensional.
The emergence of 3D hand forecasting recently advanced this field. USST \citep{H:bao2023uncertainty} is the first to address 3D hand trajectory prediction by lifting annotated 2D landmarks into 3D. EgoH4 \citep{H:hatano2025invisible} extended this paradigm by jointly predicting both 3D hand poses and trajectories using a diffusion-based model. While these methods mark important progress, they lack key properties for real-world applications, such as streaming input, language understanding, and temporal memory.
In contrast, our proposed \proposed is the first streaming 3D hand forecasting framework that integrates language instructions and a memory-augmented design, directly addressing the need for real-time, multimodal forecasting.


%% file: sec/3_method.tex
\begin{figure*}[htbp!]
	\centering
	\includegraphics[width=\linewidth]{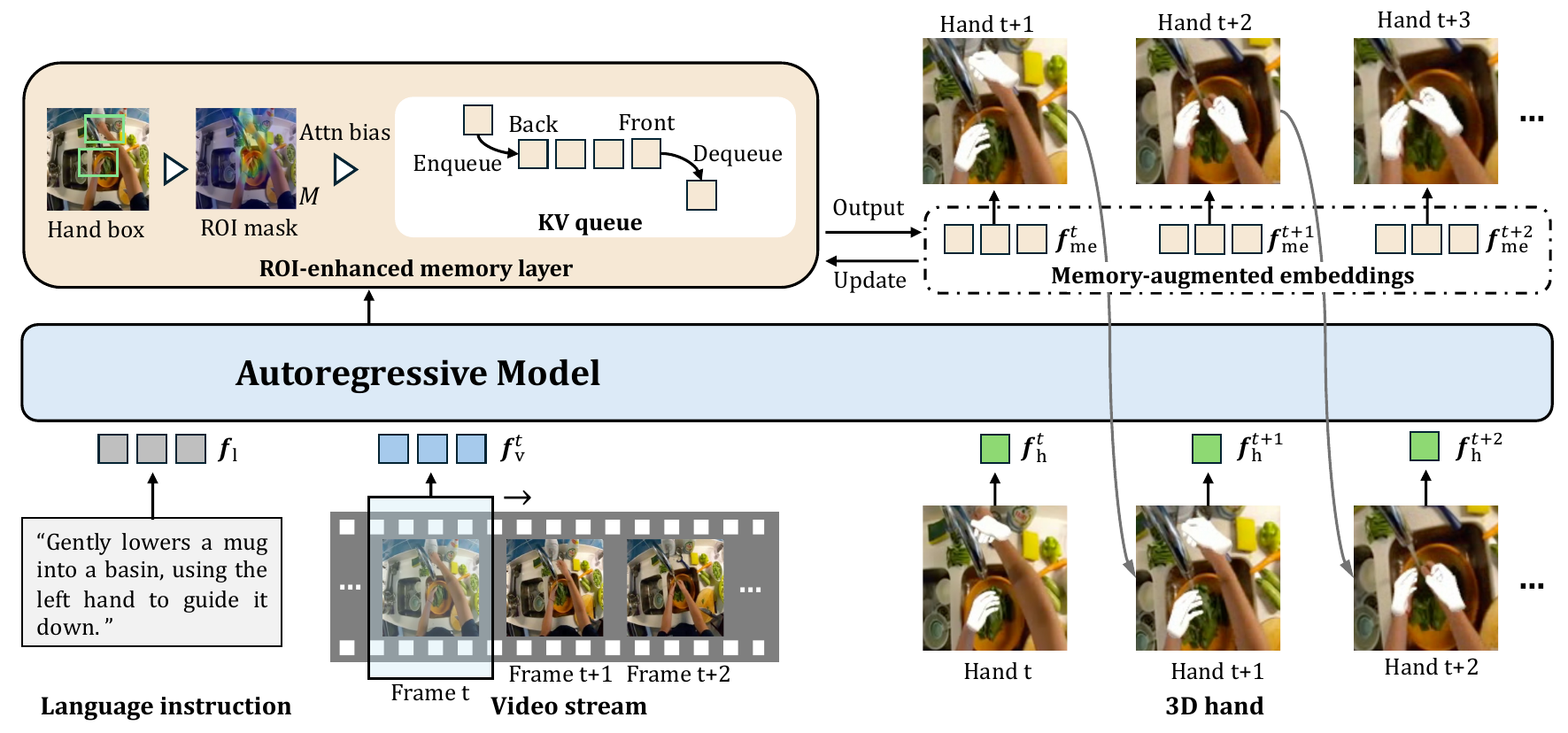}
	\vspace{-1em}
	\caption{The overview of our method. Given a streaming egocentric video, language instruction, and the current 3D hand state, our model autoregressively forecasts future 3D hand motions. The ROI-enhanced memory layer maintains a key-value queue of past embeddings, enabling temporal reasoning over streaming inputs. The ROI mask generates an attention bias that drives hand-region queries to attend more strongly to historical embeddings. The memory-augmented embeddings are then decoded to predict future hand states.}
	\label{fig:overview}
	\vspace{-1em}
\end{figure*}

\section{Method}

We formulate representation learning as autoregressive 3D hand forecasting in a streaming, multimodal setting. 
At each time step $t$, our model receives: a textual description $\mathbf{l}$ of the ongoing action, a streaming video frame $\mathbf{v}^t$ (one frame per inference step), and the current hand state $\mathbf{h}^t$ (which, during inference, is the model's own prediction from step $t-1$).
The model then outputs the predicted hand state $\mathbf{h}^{t+1}$ for the next frame. Each hand state $\mathbf{h}$ is a comprehensive representation containing the hand type (left or right), a 2D hand bounding box, the 3D hand pose represented by MANO parameters \citep{H:romero2022embodied}, and the 3D hand trajectory capturing the global hand position in space.

\subsection{Autoregressive 3D hand forecasting}\label{sec:forecast}
\cref{fig:overview} illustrates the overall workflow of our model.
At each time step $t$, given $\mathbf{l}$, $\mathbf{v}^t$, and $\mathbf{h}^t$,  three independent encoders extract textual, visual, and hand embeddings $\mathbf{f}_\text{l}$, $\mathbf{f}_\text{v}$, and $\mathbf{f}_\text{h}$.

The model's data flow is designed for streaming. The current visual and hand embeddings ($\mathbf{f}_\text{v}$, $\mathbf{f}_\text{h}$) are first processed by the ROI-enhanced memory layer $\mathcal{M}$. This layer, which is detailed in \cref{sec:memory}, maintains a key–value (KV) queue that stores past embeddings and uses this queue to refine the current visual and hand embeddings by aggregating contextual information from past entries. We then concatenate it with $\mathbf{f}_\text{l}$,
producing memory-augmented embeddings $\mathbf{f}_{\text{me}}$ that capture temporal dependencies essential for motion planning

The memory-augmented embeddings $\mathbf{f}_{\text{me}}$ are then fed into a DETR style \citep{N:carion2020end} transformer decoder $\mathcal{D}$. The decoder uses a set of learnable detection queries $\mathbf{q}_{\text{det}}$ to interact with the contextualized tokens and predict the next-frame hand states.
Formally, the structure can be described as:
\begin{equation}
\mathbf{f}_{\text{me}}^t = [\mathbf{f}_\text{l};\; \mathcal{M}(\mathbf{f}_\text{v}^t, \mathbf{f}_\text{h}^t)], \;\;\;\;
\mathbf{h}^{t+1} = \mathcal{D}(\mathbf{q}_{\text{det}}, \mathbf{f}_{\text{me}}^t),
\label{eq:forwad}
\end{equation}

\noindent where $[\ldots ;\;\ldots]$ denotes concatenation.

\subsection{ROI-enhanced memory}\label{sec:memory}
To enable effective temporal reasoning under streaming input, we design an ROI-enhanced memory layer $\mathcal{M}$. 
The memory is implemented as a fixed-size FIFO queue that stores the $N$ most recent historical embeddings, where $N$ is a hyperparameter controlling the temporal horizon. At each time step $t$, the newly encoded visual and hand embeddings are concatenated into a single vector, $\mathbf{e}^t = [\mathbf{f}_\text{v}^t; \mathbf{f}_\text{h}^t]$, which is then enqueued into the queue. This same vector $\mathbf{e}^t$ also serves as the query ($Q$) for the attention mechanism. The set of historical embeddings currently in the queue, $H = [\mathbf{e}^{t-1}; \mathbf{e}^{t-2};\ldots; \mathbf{e}^{t-N}]$, serves as both the keys ($K$) and values ($V$).

The core of our enhancement is an additive bias applied to the attention scores, using an ROI mask derived from the current 2D hand bounding box (an element of the hand state $\mathbf{h}^t$). First, we generate a binary ROI mask $M$. The $i-th$ element is assigned as $M_i=1$ if the $i-th$ visual token has corresponding image patch spatially overlaps with the 2D hand bounding box. Otherwise, $M_i=0$. This mask $M$ is then scaled by a learnable scalar $\alpha$ to create an additive bias vector $\alpha \cdot M$. This bias vector is added to the pre-softmax attention score. This operation effectively amplifies the attention originating from the hand-centric query tokens, allowing them to draw more context from the memory. The full ROI enhanced attention computation in this layer is:
\begin{equation}
\begin{aligned}
&Q = \mathbf{e}^{t} = [\mathbf{f}_\text{v}^t;\; \mathbf{f}_\text{h}^t], \;\;\;\;
K = V = H = [\mathbf{e}^{t-1};\; \mathbf{e}^{t-2};\; \ldots;\; \mathbf{e}^{t-N}], \\
&\mathcal{M}(\mathbf{e}^{t}) = Q + \sigma(\frac{Q K^{T}}{\sqrt{d}}  + \alpha M)V,
\end{aligned}
\end{equation}
where $\sigma(\cdot)$ denotes the softmax function. During training, $M$ uses the ground-truth bounding box. At inference, it is produced from the bounding box in $\mathbf{h}^t$, which is the model's own prediction.

\subsection{Training}
\noindent\textbf{Loss function.} We train \proposed in an end-to-end manner using a combination of losses that jointly supervise all components of the predicted hand state. The total loss is formulated as the weighted sum of four terms corresponding to hand type, 2D bounding box, 3D hand pose, and 3D trajectory:
\begin{equation}
\mathcal{L}=\lambda_{\text{type}}\mathcal{L}_{\text{type}} + \lambda_{\text{box}}\mathcal{L}_{\text{box}} + \lambda_{\text{pose}}\mathcal{L}_{\text{pose}} + \lambda_{\text{traj}}\mathcal{L}_{\text{traj}}.
\end{equation}
Following prior method \citep{N:carion2020end}, we perform Hungarian matching \citep{N:kuhn1955hungarian} between the predicted and ground-truth hand boxes to establish one-to-one correspondences before computing the loss.
The hand type loss $\mathcal{L}_{\text{type}}$ is a cross-entropy loss that classifies each predicted hand as left hand, right hand, or background.
The box loss is a combination of L1 loss and GIoU \citep{N:rezatofighi2019generalized} loss.
The pose loss and trajectory loss are L1 loss. Empirically, we set $\lambda_{\text{type}}=5$ and $\lambda_{\text{box}}=\lambda_{\text{pose}}=\lambda_{\text{traj}}=2$.

\noindent\textbf{Implementation details.}
We adopt the text and visual encoders from EgoHOD \citep{E:pei2025modeling}.
Specifically, the text encoder is a 12-layer GPT-like transformer \citep{N:radford2019language, N:sennrich2016neural}.
The visual encoder follows the CLIP \citep{N:radford2021learning} architecture, sharing a pretrained representation space with the text encoder.
The hand encoder is a lightweight two-layer transformer, an attention mask is applied to restrict attention to only visible hands.
The decoder in \proposed is a 4-layer transformer, followed by a MLP head, which regresses the complete hand state.

%% file: sec/4_dataset.tex
\begin{table}[t]
  \centering
  \begin{minipage}[t]{0.52\linewidth}
    \caption{Comparison of \dataset~and other egocentric datasets. \O$^*$ indicates that Ego4D contains 3D hand annotations with large errors that are not usable for learning.}
    \label{tab:dataset}
    \setlength{\tabcolsep}{.5mm}
    \centering
    \scriptsize
    \renewcommand\arraystretch{.8}
    \resizebox{\linewidth}{!}{
      \begin{tabular}{lccccc}
        \toprule[1.2pt]
        Dataset & 3D hand & Text & \#Videos & \#Frames & Open \\
        \midrule
        Ego4D \citep{E:grauman2022ego4d} & \O$^{*}$ & Coarse & 931 & $\sim$417K & Yes \\
        EgoVid \citep{E:wang2024egovid} & \O & Coarse & 5M & $\sim$120 & Yes \\
        Ego-Exo4D \citep{E:grauman2024ego} & 4.4M & Coarse & 740 & $\sim$186K & Yes \\
        EgoHOD \citep{E:pei2025modeling} & \O & Fine & 4M & $\sim$50 & Yes \\
        EgoScale \citep{E:zheng2026egoscale} & 2.25B & Fine & -- & $\sim$2,252M & No \\
        EgoLive \citep{E:li2026egolive} & 363M & Fine & 65.8K & $\sim$363M & No / Soon \\
        \dataset~(ours) & 3.95M & Fine & 247K & $\sim$90 & Yes \\
        \bottomrule[1.2pt]
    \end{tabular}}
  \end{minipage}\hfill
  \begin{minipage}[t]{0.46\linewidth}
    \centering
    \vspace{0em}
    \includegraphics[width=\linewidth]{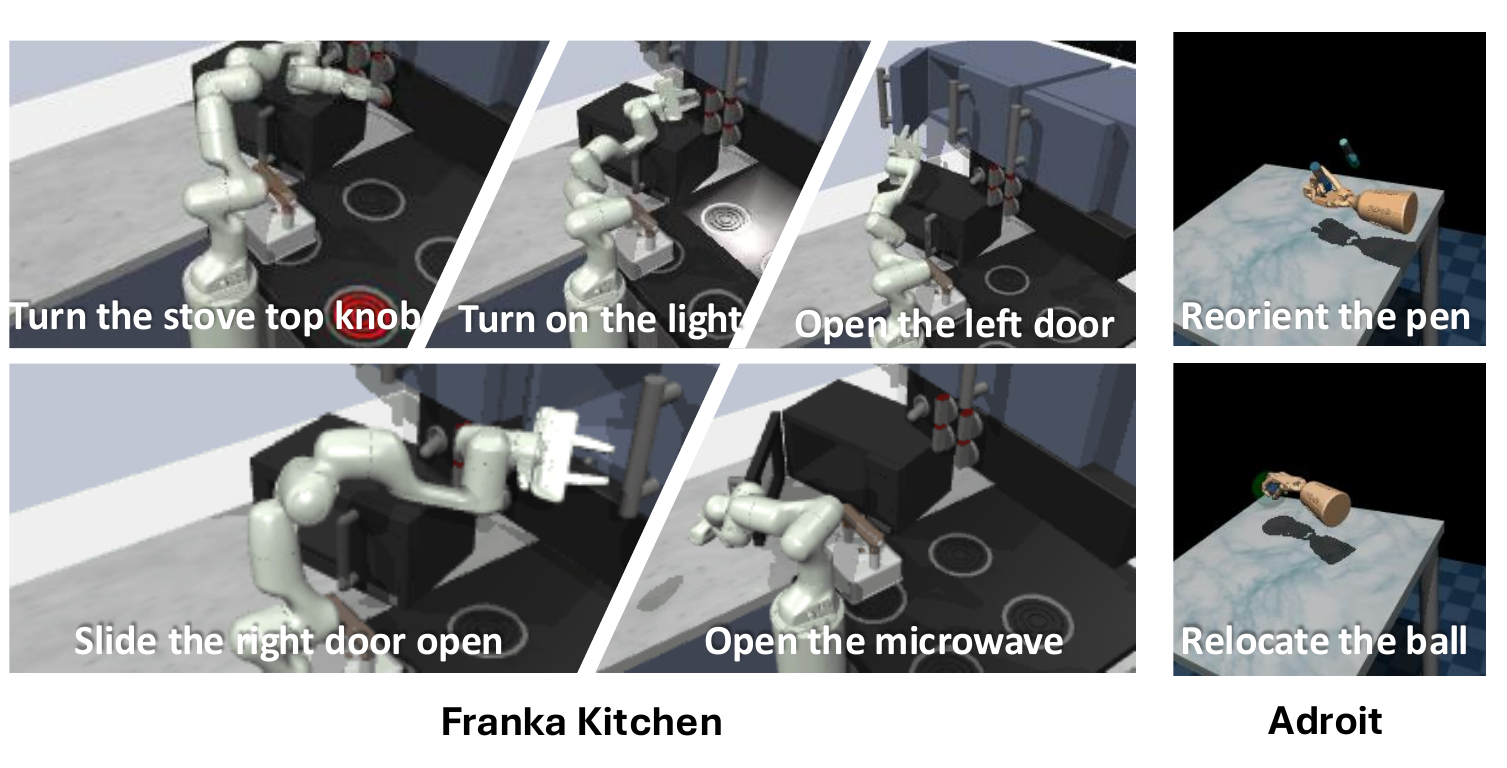}
    \captionof{figure}{Illustrations of various tasks in the Franka Kitchen \citep{R:gupta2020relay} and Adroit \citep{R:rajeswaran2017learning} simulated environments.} \label{fig:simulation}
  \end{minipage}
  \vspace{-2em}
\end{table}

\section{\dataset~dataset} \label{sec:dataset}
\textbf{Data construction.}
We construct the \dataset dataset to facilitate multimodal 3D hand forecasting with language and video input.
Our dataset is built by curating a high-quality subset of the 4M Ego4D \citep{E:grauman2022ego4d} videos. Specifically, we adopt the fine-grained, sentence-level descriptions from EgoHOD \citep{E:pei2025modeling} to provide rich textual context detailing precise hand and object movements, and we use the camera parameter annotations from EgoVid \citep{E:wang2024egovid}.

Following the video segmentation strategy of EgoHOD, we divide each video into consecutive 3-second clips, each serving as an individual training sample. Each clip's language instruction is generated solely from its content. For each clip, we employ HaMeR \citep{H:pavlakos2024reconstructing} to automatically annotate 3D hand poses at 16 frames per clip, generating MANO \citep{H:romero2022embodied} parameters and 3D hand joint positions for all visible hands. \rev{We choose HaMeR because our experiments show it provides more accurate 3D translations when focal length is specified than WiLoR \citep{H:potamias2025wilor} and HandOS \citep{H:chen2025handos}.}
To obtain physically meaningful hand trajectories, we adopt camera intrinsics from the EgoVid dataset to convert the 3D hand annotations into real-world metric coordinates. \rev{We also apply Non-Maximum Supression to remove redundant detections, followed by a carefully designed filtering strategy to discard abnormal samples, such as too large/small boxes or unrealistic depth.}

\noindent\textbf{Dataset statistics.}
\cref{tab:dataset} compares \dataset~with existing large-scale egocentric datasets.
``3D hand'' denotes the total number of 3D hand annotations; ``\#Frames'' denotes the maximum number of raw frames per clip, not all of which are annotated. Unlike prior datasets such as Ego4D \citep{E:grauman2022ego4d}, EgoVid \citep{E:wang2024egovid}, Ego-Exo4D \citep{E:grauman2024ego}, and EgoHOD \citep{E:pei2025modeling}, which either lack accurate 3D hand annotations or only provide coarse textual descriptions, \dataset~offers both high-quality 3D hand annotations and fine-grained text descriptions detailing precise hand and object movements.
EgoScale and EgoLive are larger but not  open-source yet.
In addition, our dataset is pre-segmented into short video chunks, making it suitable for forecasting tasks.
Specifically, we split \dataset into 242K clips for training and 5K for testing.
The most frequent action verbs and object categories are reported in Appendix~\ref{sec:activity-distribution}.

%% file: sec/5_experiment.tex
\begin{table}[t]
  \centering
  \begin{minipage}[t]{0.58\linewidth}
    \centering
    \caption{Evaluation on 3D hand forecasting on \dataset. SFHand$^*$ denotes the oracle variant that feeds the ground-truth hand state for autoregression.}
    \label{tab:forecast}
    \scriptsize
    \renewcommand\arraystretch{.75}
    \setlength{\tabcolsep}{0.5mm}
    \resizebox{\linewidth}{!}{%
      \begin{tabular}{@{}lccccccc@{}}
        \toprule[1.2pt]
        \multirow{2}{*}{Method} & \multirow{2}{*}{Streaming} & \multirow{2}{*}{Instruct} & \multicolumn{2}{c}{Trajectory} & \multicolumn{2}{c}{Hand Pose} & \multirow{2}{*}{FPS} \\
        && & ADE & FDE & JPE & PA-JPE \\
        \midrule
        Static & \wrong & \wrong & 26.71 & 29.40 & 5.92 & 1.44 & - \\
        USST \citep{H:bao2023uncertainty} & \wrong & \wrong & 49.15 & 55.20  & - & - & 0.66\\
        EgoH4 \citep{H:hatano2025invisible} & \wrong & \wrong & 22.56 & 22.87 & 5.75 & 2.30 & 0.25\\
        WiLoR \citep{H:potamias2025wilor} & \correct & \wrong & 22.30 & 21.66 & \underline{3.43} & \textbf{0.87} & 59.7\\
        HaMeR \citep{H:pavlakos2024reconstructing} & \correct & \wrong & 19.69 & 19.10 & 3.51 & \underline{0.92} & \textbf{105.1}\\
        HaMeR + Instruct & \correct & \correct & \underline{18.03} & \underline{17.13} & 3.55 & 0.93 & \underline{63.9}\\
        EgoHOD \citep{E:pei2025modeling} & \correct & \correct & 19.58 & 19.07 & 3.65 & 0.93 & 55.3\\
        \midrule
        \proposed & \correct & \correct & \textbf{12.65} & \textbf{13.08} & \textbf{3.38} & \underline{0.92} & 33.4\\
        \proposed{}$^*$ & \correct & \correct & 10.39 & 9.74 & 2.91 & 0.79 & 58.8\\
        \bottomrule[1.2pt]
    \end{tabular}}
  \end{minipage}\hfill
  \begin{minipage}[t]{0.4\linewidth}
    \centering
    \caption{Cross-dataset evaluation on GigaHands. All evaluations are conducted under a zero-shot setting.}
    \label{tab:gigahands}
    \scriptsize
    \renewcommand\arraystretch{1.1}
    \setlength{\tabcolsep}{0.8mm}
    \resizebox{\linewidth}{!}{%
      \begin{tabular}{@{}lcc|cc@{}}
        \toprule[1.2pt]
        \multirow{2}{*}{Method} & \multicolumn{2}{c}{Trajectory} & \multicolumn{2}{c}{Hand Pose} \\
        & ADE & FDE & JPE & PA-JPE \\
        \midrule
        HaMeR \citep{H:pavlakos2024reconstructing} & 16.00 & 21.31 & \underline{4.74} & 1.30\\
        WiLoR \citep{H:potamias2025wilor} & \underline{15.96} & \underline{17.64} & 5.79 & \textbf{1.09}\\
        HaMeR + Instruct & 17.84 & 21.74 & 5.28 & 1.24\\
        EgoHOD \citep{E:pei2025modeling} & 17.40 & 18.28 & 6.47 & 1.29\\
        \midrule
        \proposed & \textbf{12.99} & \textbf{15.23} & \textbf{4.25} & \underline{1.22}\\
        \bottomrule[1.2pt]
    \end{tabular}}
  \end{minipage}
  \vspace{-1em}
\end{table}

\section{Experiment}

\subsection{Experimental setup}
We evaluate both the \textbf{3D hand forecasting} and \textbf{representation learning}. The forecasting accuracy is evaluated on the test split of \dataset, while the learned representations is evaluated on downstream \textbf{robotic manipulation}. We use two standard simulation-based benchmarks: Franka Kitchen environment \citep{R:gupta2020relay} for gripper-based manipulation, and Adroit environment \citep{R:rajeswaran2017learning} for dexterous hand control (\cref{fig:simulation}).

We pre-train \proposed the model on our \dataset dataset. We use the AdamW optimizer with an initial learning rate of 2e-4, and train on 8 GPUs with a batch size of 256 per GPU. For the manipulation experiments, we follow the established practice~\citep{R:nair2023r3m,R:karamcheti2023language,R:radosavovic2023real}. We freeze the pre-trained encoders and attach a lightweight, task-specific prediction head on top of the extracted representations. This head is then trained to map the encoded features to the corresponding robotic control.
\rev{To mitigate exposure bias, we follow standard practices by applying data augmentation, including random noise and dropout, to the input hand parameters during training.}


\subsection{Evaluation on 3D hand forecasting}\label{sec:eval-forecast}
We report comparisons the following metrics in centimeters (cm). We measure trajectory error using \textbf{ADE} (Average Displacement Error), the mean Euclidean distance over all forecasted frames, and \textbf{FDE} (Final Displacement Error), the distance at the final frame. We measure pose error using \textbf{JPE} (Joint Position Error), the wrist-aligned mean per-joint error, and \textbf{PA-JPE} (Procrustes-Aligned Joint Position Error) \citep{N:gower1975generalized}. We also report the \textbf{FPS} of each method. All tests are conducted on a single NVIDIA L40S GPU with a batch size of 1 and an input resolution of 224×224, maintaining a 3-second memory horizon.

We construct a ``Static'' baseline, where we uses the first observed frame as the prediction for all future frames. Comparison methods contain: \textit{USST} \citep{H:bao2023uncertainty}, \textit{EgoH4} \citep{H:hatano2025invisible}, \textit{HaMeR} \citep{H:pavlakos2024reconstructing}, \textit{WiLoR} \citep{H:potamias2025wilor}, and \textit{EgoHOD} \citep{E:pei2025modeling}. We also build an oracle variant, \proposed*, which feeds the ground-truth hand states for forecasting.
As shown in \cref{tab:forecast}, \proposed substantially outperforms all prior approaches across all metrics, indicating that it effectively learns future hand-object interaction dynamics. Notably, non-streaming methods like USST and EgoH4 exhibit both lower accuracy and very low FPS, as they require accumulating full video sequences. This confirms their unsuitability for real-time tasks. Compared to the HaMeR and WiLoR baselines, \proposed still achieves significant and consistent accuracy gains, demonstrating the clear benefits of integrating multimodal inputs and temporal reasoning. While HaMeR is faster (over 100 FPS) due to its visual-only encoder, our \proposed model runs at 33.4 FPS. This speed, which accounts for our multimodal encoders and autoregressive inference, remains sufficient for real-time applications. The oracle variant, \proposed{}*, achieves 58.8 FPS. These results confirm that \proposed sets a state-of-the-art in forecasting accuracy while being efficient for real-time use.
A component-wise runtime breakdown measured separately on an RTX 5090 is provided in Appendix~\ref{sec:runtime-breakdown}.

To evaluate \proposed's generalizability to real world, we conduct zero-shot evaluations on GigaHands dataset \citep{H:fu2025gigahands}, a recent dataset providing real 3D hand annotations from a
multi-camera capture setup. Specifically, we select camera views from the GigaHands demo data that closely align with an egocentric perspective (within $\pm 60^\circ$ of the egocentric optical axis \citep{I:lin2002effects}) to form our test set. As presented in \cref{tab:gigahands}, \proposed reduces ADE by 18.6\% and JPE by 10.3\% over the strongest per-frame estimators (WiLoR and HaMeR, respectively). These results confirm that \proposed effectively learns transferable dynamics of future hand-object interactions. Such transferable representation is exactly what downstream embodied control demands.

\subsection{Ablation on ROI-enhanced memory}\label{sec:roi-ablation}

\begin{table}[t]
  \centering
  \begin{minipage}[t]{0.4\linewidth}
    \vspace{0pt}
    \centering
    \caption{Ablation study on the ROI-enhanced memory layer. ``Recall.5'' denotes recall at 0.5 IoU.}
    \label{tab:memo-ablation}
    \scriptsize
    \renewcommand\arraystretch{1.2}
    \setlength{\tabcolsep}{2pt}
    \resizebox{\linewidth}{!}{%
      \begin{tabular}{@{}cc|ccccc@{}}
        \toprule[1.2pt]
        \multirow{2}{*}{Memory} & \multirow{2}{*}{ROI} & \multicolumn{2}{c}{Trajectory} & \multicolumn{2}{c}{Hand Pose} & \multirow{2}{*}{Recall.5} \\
        && ADE & FDE & JPE & PA-JPE & \\
        \midrule
        \wrong & \wrong & 15.56 & 18.38 & \textbf{3.29} & \textbf{0.92} & \textbf{0.72} \\
        \correct & \wrong & 18.34 & 22.23 & 3.36 & \textbf{0.92} & 0.71 \\
        \correct & \correct & \textbf{12.65} & \textbf{13.08} & 3.38 & \textbf{0.92} & 0.70 \\
        \bottomrule[1.2pt]
    \end{tabular}}
  \end{minipage}\hfill
  \begin{minipage}[t]{0.58\linewidth}
    \vspace{0pt}
    \centering
    \includegraphics[width=\linewidth]{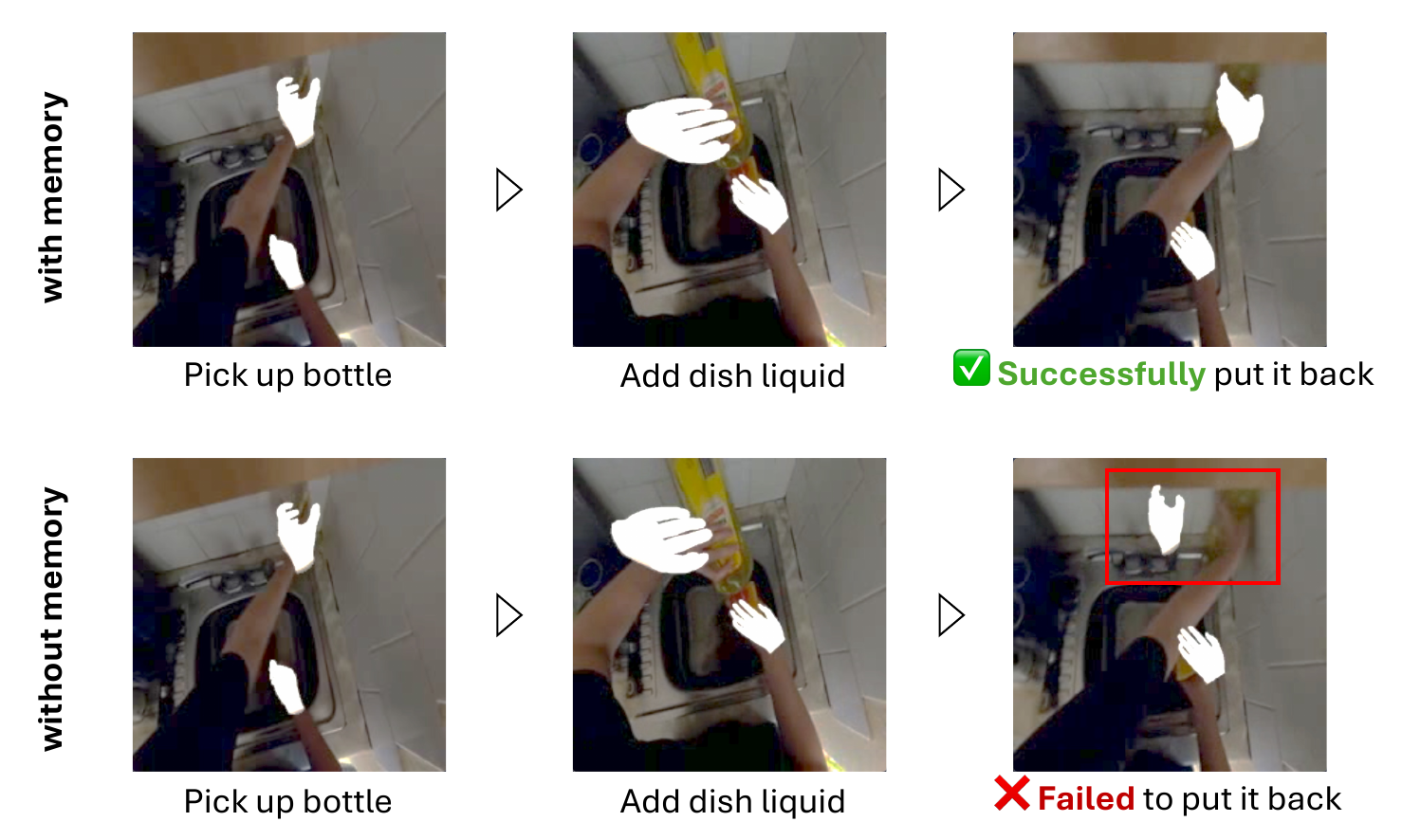}
    \captionof{figure}{Function of memory layer. All hands are forecasted from previous video frames and hands.}
    \label{fig:memo_func}
  \end{minipage}
  \vspace{-2em}
\end{table}

We further examine the contribution of the proposed ROI-enhanced memory layer through ablation experiments, as summarized in \cref{tab:memo-ablation}.
Three configurations are compared: 1) without memory, 2) with the memory layer but without ROI enhancement, and 3) with the full ROI-enhanced memory.
The results show that introducing the ROI-enhanced memory layer improves overall 3D forecasting performance, particularly in trajectory metrics (ADE and FDE). This demonstrates that maintaining a temporal buffer of past embeddings helps the model reason over temporal dependencies and anticipate future motions more accurately.
Meanwhile, the vanilla memory without ROI enhancement performs even worse than the no-memory baseline.
We attribute this to the fact that a naive attention mechanism, without guidance, allows noisy or irrelevant parts of the current input (e.g., background visual tokens) to retrieve information from the memory.

\noindent\textbf{Qualitative results.} \cref{fig:memo_func} presents qualitative comparisons between models with and without the proposed memory layer. In this example, the task involves picking up a bottle, adding dish liquid, and then returning the bottle to its original position.
The model with memory successfully predicts the correct future motion, accurately anticipating the put-back location after pouring.
In contrast, the model without memory fails to infer the final motion, predicting an incorrect hand position when the bottle should be returned. This failure arises because the model lacks access to historical context and therefore cannot infer the original place of the bottle.
These results clearly illustrate that the memory layer enables temporal reasoning for motion planning.

Additional ablations on input modalities and forecasting horizon are provided in Appendix~\ref{sec:ablation}. Visualizations are presented in Appendix~\ref{sec:qualitative}.

\begin{table*}[t]
\caption{Results of robotic manipulation on Franka kitchen. We report the success rate (\%) over 50 randomly sampled trajectories. The best result is \textbf{bolded}, and the second-best is \underline{underlined}. ``INSUP.'' denotes classification-based supervised pretraining on ImageNet.}
\vspace{-.5em}
\label{tab:franka}
\centering
\small
\renewcommand\arraystretch{.9}
\resizebox{\linewidth}{!}{
\begin{tabular}{l|cc|ccccc|c}
\toprule[1.2pt]
Method & Backbone & Param. & Flip switch & Open microwave & Slide door & Turn knob & Open door & Average\\
\midrule
INSUP. \citep{N:he2016deep} & ResNet50 & 25.6M & 50.0 & 26.7 & 75.7 & 28.0 & 18.0 & 39.7 \\
CLIP \citep{N:radford2021learning}  & ResNet50 & 25.6M & 41.7 & 24.7 & 86.3 & 26.3 & 13.0 & 38.4 \\
MVP \citep{R:radosavovic2023real} & ViT-Base & 86M & 90.7 & 41.0 & \textbf{100.0} & 83.3 & 50.3 & 76.5 \\
Voltron \citep{R:karamcheti2023language} & ViT-Base & 86M & 91.0 & 41.0 & \underline{99.3} & 76.0 & 45.3 & 70.5 \\
MPI \citep{R:zeng2024learning} & ViT-Base & 86M & \underline{93.7} & \underline{54.0} & \textbf{100.0} & \textbf{89.0} & \textbf{57.7} & 78.9 \\
Ours & ViT-Base & 86M & \textbf{97.7} & \textbf{58.0} & \textbf{100.0} & \underline{88.0} & \underline{55.7} & \textbf{79.9}\\
\bottomrule[1.2pt]
\end{tabular}}
\vspace{-1em}
\end{table*}

\begin{figure}[t]
\centering
\begin{minipage}{0.58\linewidth}
\centering
\includegraphics[width=\linewidth]{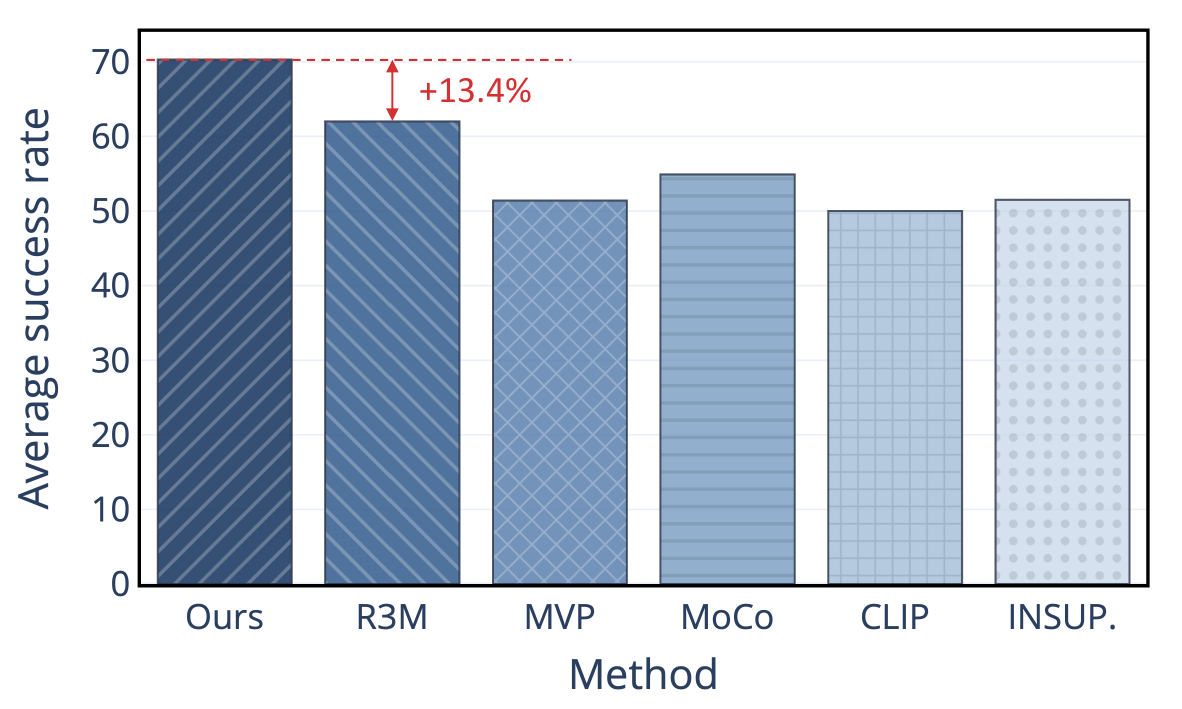}
\caption{Results of manipulation on Adroit. We report the success rate (\%) over 50 randomly sampled trajectories. ``INSUP.'' denotes classification-based supervised pretraining on ImageNet.}
\label{fig:adroit}
\end{minipage}
\hfill
\begin{minipage}{0.38\linewidth}
\centering
\includegraphics[width=\linewidth]{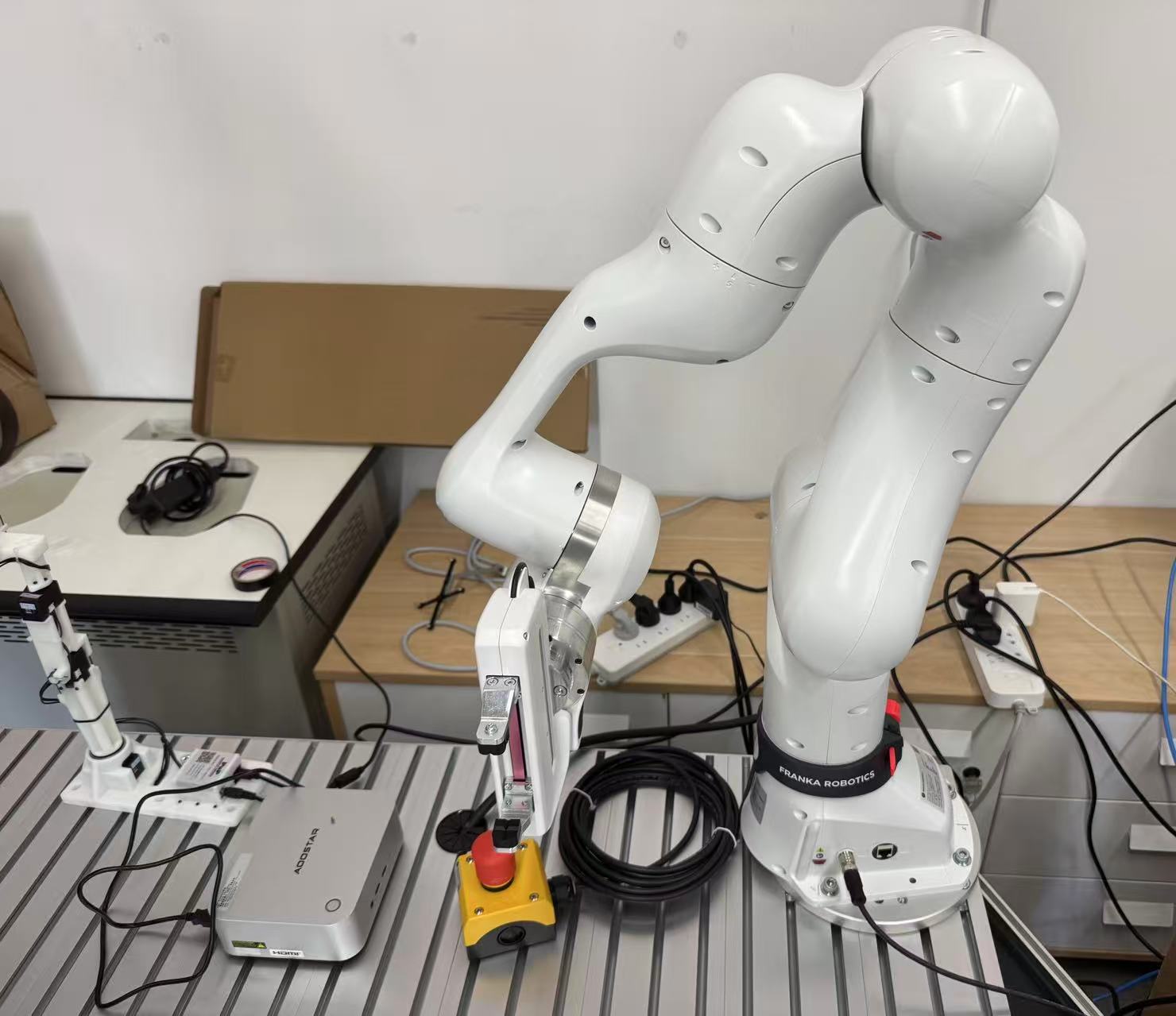}
\caption{Our real-world evaluation setting with a Franka Research 3 (FR3, v2.1), a 7-DoF arm with a parallel-jaw gripper.}
\label{fig:real_robot}
\end{minipage}
\vspace{-2em}
\end{figure}

\subsection{Evaluation on embodied manipulation in simulation environment}
\textbf{Evaluation Details.} We evaluate the learned representations by transferring them to the Franka Kitchen and Adroit environments.
Franka Kitchen includes 5 distinct manipulation tasks, each observed from 2 camera viewpoints.
Adroit contains 2 dexterous hand manipulation tasks, each observed from 3 camera viewpoints.
The control policy receives vision–language representations extracted from our pre-trained model, combined with proprioceptive states (\ie, joint velocities), as input.
Following prior work \citep{R:zeng2024learning,R:nair2023r3m}, we incorporate contrastive learning and frame reconstruction objectives during pre-training to preserve dense visual information.
We train a separate policy head for each task, which imitates RL experts \citep{R:zeng2024learning}.
Evaluation follows the protocol of \citep{R:karamcheti2023language,R:nair2023r3m}: computing the average success rate across all tasks, viewpoints, and 3 random seeds.

\noindent \textbf{Franka Kitchen.} As shown in \cref{tab:franka}, representation learning frameworks tailored for robotic manipulation (MVP, Voltron, MPI) demonstrate a clear advantage over conventional visual pre-training approaches widely used in computer vision, such as ImageNet classification pre-training \citep{N:he2016deep} and CLIP \citep{N:radford2021learning}.
All of these baselines pretrain with purely perceptual objectives (\eg, MAE and contrastive learning), and therefore do not explicitly model the next motion.
In contrast, our predictive representation learning surpasses all of them with a 79.9\% average success rate, setting a new state of the art on the Franka Kitchen benchmark.
Notably, \proposed shares the same ViT-Base backbone and parameter budget as MVP, Voltron, and MPI. The gap therefore stems entirely from the pretraining objective, not from added model capacity.

\noindent\textbf{Adroit.} \cref{fig:adroit} presents the average success rates on the Adroit benchmark.
Compared with a range of baselines, our method achieves the highest performance, outperforming the previous best model (R3M) by 13.4\% in average success rate.
The largest gains occur on Adroit, whose dexterous hand morphology is closer to human hands.

Overall, transfer gains vary across embodiments and tasks: the improvement on Franka Kitchen is modest (1.0\% over MPI), whereas the gains on Adroit are substantially larger. However, the benefit is not confined to dexterous hands: \proposed also outperforms MPI by 10.0\% on a real-world Franka Research 3 parallel-jaw gripper (\cref{tab:real_robot}). These results support the transferability of hand-forecasting representations without implying uniform gains across embodiments.

\vspace{-.5em}
\subsection{Real-robot validation}
\label{sec:real_robot}
\vspace{-.5em}

To validate that our representation transfers beyond simulation, we deploy SFHand to a physical Franka Research 3 (as shown in \cref{fig:real_robot}), a 7-DoF arm with a parallel-jaw gripper, and compare against MPI~\cite{R:zeng2024learning}, the strongest baseline in our simulation experiments. Both methods use a ViT-Base backbone and an identical frozen-encoder + MLP policy head trained via behavior cloning~\cite{R:nair2023r3m, R:radosavovic2023real, R:zeng2024learning}.
We also evaluate the pretrained EgoHOD encoder~\citep{E:pei2025modeling} directly to assess the benefit of forecasting beyond its prior egocentric pretraining.
We design five tasks covering grasping (pick-and-place a cup), articulated-object interaction (open a drawer, close a microwave door), precise contact (flip a light switch), and tool use (pour from a bottle). For each task, we collect 30 teleoperation demonstrations, train policies with 3 random seeds, and evaluate over 10 rollouts per seed with randomized initial object placement. Success criteria are fixed before evaluation begins.

As shown in Table~\ref{tab:real_robot}, SFHand achieves an average success rate of 72.0\%, compared with 62.0\% for MPI and 65.3\% for EgoHOD. Since SFHand is initialized from EgoHOD, the 6.7\% gain over EgoHOD supports the benefit of forecasting pretraining beyond the initial egocentric representation. Relative to MPI, the gains are largest on pouring (+20.0\%) and switching (+13.3\%), while performance is comparable on simple grasping. These results confirm our central clam: forecasting 3D hand motion is a stronger pretraining objective than perception-only objectives when transferring to physical robot deployment.

\begin{table}[t]
\centering
\caption{Mean success rates (\%) on a real-world Franka Research 3 arm over 3 seeds, with 10 rollouts per seed.}
\label{tab:real_robot}
\small
\renewcommand\arraystretch{.8}
\begin{tabular}{l|ccccc|c}
\toprule
Method & Pick-place & Drawer & Microwave & Switch & Pour & Avg. \\
\midrule
MPI~\cite{R:zeng2024learning} & 83.3 & 70.0 & 63.3 & 50.0 & 43.3 & 62.0 \\
EgoHOD~\citep{E:pei2025modeling} & 83.3 & 73.3 & 66.7 & 53.3 & 50.0 & 65.3 \\
Ours & \textbf{86.7} & \textbf{76.7} & \textbf{70.0} & \textbf{63.3} & \textbf{63.3} & \textbf{72.0} \\
\bottomrule
\end{tabular}
\vspace{-2em}
\end{table}

%% file: sec/6_conclusion.tex
\vspace{-.5em}
\section{Conclusion}
\vspace{-.5em}
In this work, we introduce \proposed, a representation learning method via streaming egocantric 3D hand forecasting. Our model is equipped with an ROI-enhanced memory layer that enables temporal reasoning, and an autoregressive architecture that brings instruction-following capability.
To support this task, we construct \dataset, the first open-source large-scale dataset providing synchronized egocentric videos, fine-grained language descriptions, and accurate 3D hand annotations.
Experiments demonstrate that \proposed~achieves state-of-the-art accuracy in 3D hand forecasting. Further, its predictive representations effectively transfer to downstream embodied manipulations and show leading performance.

\noindent\textbf{Limitation.} While SFHand demonstrates strong performance, its current implementation relies on MANO-based hand representations, which may limit generalization to hands with extreme articulations or occlusions. Instruction's semantic ambiguity is another limitation. We analyze the instruction using word clouds and find that fine-grained words like ``brush'' and ``steady'' predominate in low-error samples, whereas ambiguous terms like ``place'' are common in high-error cases.
Transfer gains also vary across embodiments and tasks: they are largest on the human-hand-like Adroit benchmark and modest on Franka Kitchen, although real-world parallel-jaw gripper tasks also benefit. Our results therefore do not establish uniform gains across embodiments.
Autoregressive forecasting is also susceptible to error accumulation: feeding predicted rather than ground-truth hand states increases ADE from 10.39 to 12.65 cm (\cref{tab:forecast}). This gap indicates sensitivity to errors in previous predictions, which may compound over longer forecasting horizons. The impact of this drift on long-horizon downstream manipulation remains to be quantified.

\section*{Acknowledgments}
This work was supported by JST ASPIRE (Grant Number JPMJAP2303), JSPS Grant-in-Aid for Scientific Research (B) (Grant Number JP24K02956), and Grant-in-Aid for JSPS Fellows (Project Number JP25KJ1180).

%% file: sec/X_suppl.tex
\clearpage
\appendix
\section*{Appendix}

\section{Ablation study on 3D hand forecasting}\label{sec:ablation}

\subsection{Ablation on input modalities}

To investigate the contribution of each input modality, we conduct an ablation study over video, text, and hand inputs, as summarized in \cref{tab:modal-ablation}.
This experiment is measured using both trajectory metrics, hand pose metrics, and 2D box Recall@0.5 IoU.
Notably, when the hand input is removed, the model degenerates from autoregressive to regressive, since it no longer receives past predictions as context.

\noindent \textbf{Results.} The results reveal clear trends across modalities. Removing the hand input causes a substantial performance drop in trajectory (ADE/FDE) metrics, confirming the importance of previous hand information for accurate 3D motion forecasting.
Excluding language input degrades performance in trajectory but shows less impact in hand pose. Such a result stems from the different information they depend on: Trajectory reflects high-level intent, which is directly guided by language instructions, whereas hand pose is primarily guided by physical constraints (object affordance) that text cannot easily finetune.

As can be observed, the video modality contributes the most to 3D forecasting accuracy, as visual cues capture scene dynamics, object interactions, and hand–object relationships that are crucial for predicting realistic future motions.
Video also plays the most important role for 2D bounding box prediction, yielding the lowest Recall@0.5 IoU when removed, since it preserves rich spatial context and visual detail necessary for precise localization.

\begin{table}[htbp!]
  \caption{Ablation on input modalities. ``Recall.5'' indicates recall at 0.5 IoU threshold.}
  \label{tab:modal-ablation}
  \centering
  \scriptsize
  \renewcommand\arraystretch{.7}
  \setlength{\tabcolsep}{3mm}
  \begin{tabular}{@{}ccc|ccccc@{}}
    \toprule[1.2pt]
    \multirow{2}{*}{Video} & \multirow{2}{*}{Text} & \multirow{2}{*}{Hand} & \multicolumn{2}{c}{Trajectory} & \multicolumn{2}{c}{Hand Pose} & \multirow{2}{*}{Recall.5} \\
    && & ADE & FDE & JPE & PA-JPE & \\
    \midrule
    \correct & \wrong & \wrong & 21.09 & 20.48 & 3.39 & 0.93 & \textbf{0.72} \\
    \correct & \correct & \wrong & 19.51 & 18.47 & 3.55 & 0.93 & \textbf{0.72} \\
    \correct & \wrong & \correct & 14.40 & 15.47 & \textbf{3.38} & \textbf{0.92} & \textbf{0.72} \\
    \wrong & \correct & \correct & 18.10 & 22.67 & 4.25 & 1.00 & 0.40\\
    \correct & \correct & \correct & \textbf{12.65} & \textbf{13.08} & \textbf{3.38} & \textbf{0.92} & 0.70 \\
    \bottomrule[1.2pt]
  \end{tabular}
\end{table}



\begin{figure}[htbp!]
  \centering
  \includegraphics[width=0.6\linewidth]{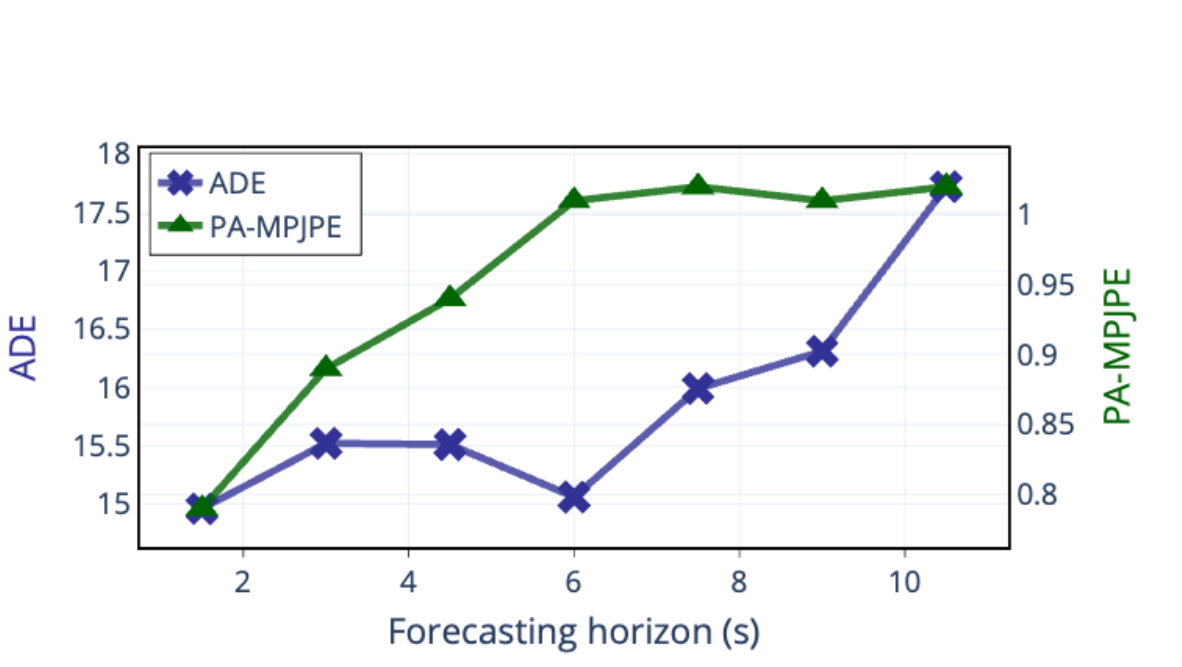}
  \caption{Forecasting accuracy with increasing forecasting horizons.}
  \label{fig:horizon}
\end{figure}

\rev{
  \subsection{Analysis of forecasting horizon}
We conduct experiments on the GigaHands dataset \citep{H:fu2025gigahands} with forecasting horizons ranging from 1.5 to 10.5 seconds. As shown in \cref{fig:horizon}, both trajectory (ADE) and hand pose (PA-MPJPE) errors correlate positively with the forecasting horizon. This upward trend reflects the naturally increasing difficulty of long-term prediction. The overall error growth remains within a predictable range, demonstrating that our model's robustness even as the forecasting becomes significantly more challenging.}

\begin{figure*}[t]
  \centering
  \includegraphics[width=\linewidth]{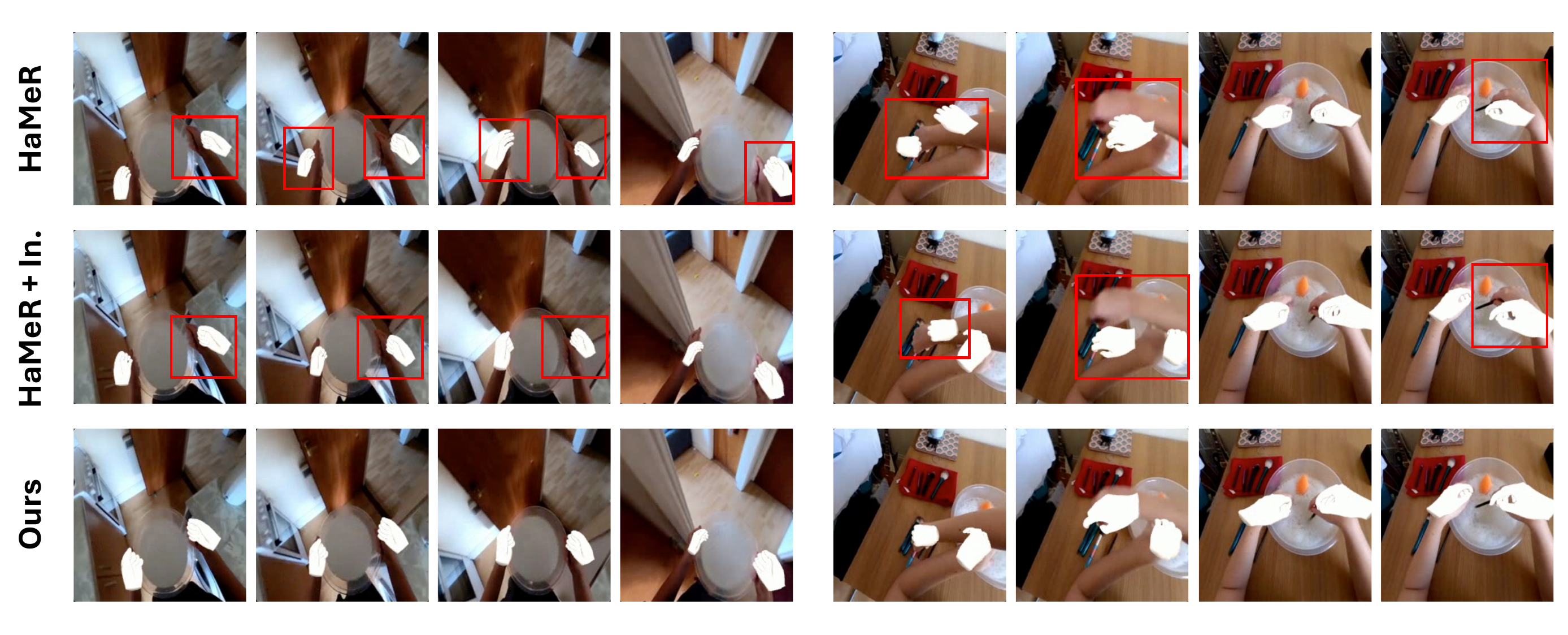}
  \caption{Qualitative comparison between our method and HaMeR. ``HaMeR + In.'' indicates HaMeR incorporating language instructions for forecasting. \textcolor{red}{Red rectangles} highlight incorrect hand positions or hand poses. All hands are forecasted from previous video frames and hand states.}
  \label{fig:visualize}
\end{figure*}

\section{Qualitative result} \label{sec:qualitative}
\textbf{Motivation.} To further demonstrate the effectiveness of our method, we present qualitative comparisons with HaMeR \citep{H:pavlakos2024reconstructing}, the best-performing previous 3D hand forecasting approach in \cref{sec:eval-forecast}, as shown in \cref{fig:visualize}. This experiment aims to visualize the differences in spatial accuracy and temporal consistency between the two models.

\noindent\textbf{Results.} As illustrated in \cref{fig:visualize}, HaMeR often produces inaccurate hand positions and poses, highlighted by red rectangles. These errors typically occur when the model loses temporal coherence across frames, causing unnatural hand drift, misalignment with manipulated objects, or implausible inter-hand interactions. Since HaMeR lacks explicit temporal memory and multimodal reasoning, it struggles to maintain consistent motion across extended forecasting horizons.
In contrast, our \proposed~model generates stable and plausible predictions across consecutive frames. The hands remain well-aligned with objects and exhibit smooth motion transitions, demonstrating a more accurate understanding of ongoing tasks.

These qualitative results clearly show that \proposed achieves superior temporal consistency and spatial alignment compared to prior methods. By combining memory-augmented temporal reasoning with multimodal (video, text, and hand) understanding, our model produces more coherent and physically grounded 3D hand forecasts.

\section{Activity distribution in \dataset}\label{sec:activity-distribution}
\cref{tab:activity-distribution} reports the 14 most frequent action verbs and object categories in \dataset, together with their percentages. These categories span activities such as placing, taking, cutting, washing, and inspecting, and include objects such as containers, tools, and plants.

\begin{table}[t]
  \centering
  \caption{The 14 most frequent action verbs and object categories in \dataset. The table shows only the most frequent categories, not the full distribution.}
  \label{tab:activity-distribution}
  \small
  \setlength{\tabcolsep}{3pt}
  \renewcommand{\arraystretch}{1.0}
  \begin{tabular}{rlr rlr @{\hspace{10pt}} rlr rlr}
    \toprule
    \multicolumn{6}{c}{\textbf{Action verb}} & \multicolumn{6}{c}{\textbf{Object}} \\
    \cmidrule(r){1-6}\cmidrule(l){7-12}
    \# & Class & \% & \# & Class & \% & \# & Class & \% & \# & Class & \% \\
    \midrule
    1 & put    & 18.9 &  8 & turn    & 2.4 & 1 & hand      & 5.1 &  8 & book   & 2.3 \\
    2 & take   & 16.5 &  9 & remove  & 2.3 & 2 & paper     & 4.6 &  9 & phone  & 2.0 \\
    3 & move   &  5.9 & 10 & wash    & 2.2 & 3 & container & 3.9 & 10 & chip   & 1.9 \\
    4 & hold   &  5.3 & 11 & touch   & 2.2 & 4 & knife     & 3.1 & 11 & fork   & 1.6 \\
    5 & cut    &  2.8 & 12 & wipe    & 2.1 & 5 & cloth     & 2.7 & 12 & faucet & 1.5 \\
    6 & open   &  2.5 & 13 & walk    & 2.0 & 6 & plant     & 2.6 & 13 & bag    & 1.4 \\
    7 & adjust &  2.5 & 14 & inspect & 2.0 & 7 & card      & 2.4 & 14 & water  & 1.3 \\
    \bottomrule
  \end{tabular}
\end{table}

\section{Runtime breakdown}\label{sec:runtime-breakdown}
\cref{tab:runtime-breakdown} reports the component-wise inference time of HaMeR, EgoHOD, and \proposed in milliseconds. All measurements in this table were obtained on a single NVIDIA RTX 5090 GPU, and therefore should not be compared directly with the L40S-based FPS measurements in \cref{tab:forecast}.

The memory attention itself is lightweight, requiring only 0.25 ms. The main overhead comes from ROI inference (5.18 ms), which obtains the hand bounding box through an extra forward pass and nearly doubles the total runtime. Thus, the additional cost is primarily due to ROI inference rather than attention over cached features.

\begin{table}[t]
  \centering
  \caption{Component-wise inference time (ms) on a single NVIDIA RTX 5090 GPU. A dash indicates that the component is not used.}
  \label{tab:runtime-breakdown}
  \small
  \renewcommand{\arraystretch}{1.0}
  \begin{tabular}{l|cccccc|c}
    \toprule
    Method & Text enc. & Visual enc. & Hand enc. & Memory & ROI infer. & Decoder & Total \\
    \midrule
    HaMeR & -- & 1.98 & -- & -- & -- & 1.08 & 3.06 \\
    EgoHOD & 1.86 & 1.96 & -- & -- & -- & 1.07 & 4.89 \\
    \proposed & 1.88 & 2.01 & 0.55 & 0.25 & 5.18 & 0.92 & 10.79 \\
    \bottomrule
  \end{tabular}
\end{table}